# A Greedy, Flexible Algorithm to Learn an Optimal Bayesian Network Structure

**Amir Arsalan Soltani**

Department of Computer Science and Engineering

University at Buffalo, The State University of New York

Buffalo, NY, 14260

amirarsa@buffalo.edu

**Abstract**

In this report paper we first present a report of the Advanced Machine Learning Course Project on the provided data set and then present a novel heuristic algorithm for exact Bayesian network (BN) structure discovery that uses decomposable scoring functions. Our algorithm follows a different approach to solve the problem of BN structure discovery than the previously used methods such as Dynamic Programming (DP) and Branch and Bound to reduce the search space and find the global optima space for the problem. The algorithm we propose has some degree of flexibility that can make it more or less greedy. The more the algorithm is set to be greedy, the more the speed of the algorithm will be, and the less optimal the final structure. Our algorithm runs in a much less time than the previously known methods and guarantees to have an optimality of close to 99%. Therefore, it sacrifices less than one percent of score of an optimal structure in order to gain a much lower running time and make the algorithm feasible for large data sets (we may note that we never used any toolbox except for result validation)

## 1      Introduction

Modeling handwriting data and understand the ways it evolves, in general, is an important task currently being persuaded by forensic researchers in the world. One Machine learning provides great tools to the researchers to unfold the data and export meaningful information from it. One of the most important tools is Bayesian networks which captures a compact and meaningful representation of statistical relationships between the features in any data set. We use Bayesian networks to the model the provided children hand writing data and run some inference tasks on it.

Bayesian network (BN) is a directed acyclic graph (DAG) which is a compact way of modeling the statistical relationship between the variables by having parent-child dependency/relationship among them. As the amount of available digital data is growing overwhelmingly, the necessity of modeling data is becoming more important every day, specially for data sets having a large number of variables. Bayesian networks (BNs) are one of the most powerful tools currently available to model different data sets with different number of variables. Therefore, Bayesian network structure discovery, which is equivalent to find statistical relationship of variables, is a crucial task since the better the structure the higher the quality of inference and knowledge discovery will be.

On 2003 the problem of learning a BN structure with more than two parents per node was proved to be NP-Hard [2]. However, researchers have pushed the boundaries forward by adopting different techniques such as Dynamic Programming and other heuristics to reduce the search space to a more optimal search space and find the best set of parents for each variable using decomposable scoring functions such as BIC, AIC and different variants of BD. Also, since different Bayesian network may have equal scores [7], a set of variables may not have only one optimal Bayesian network structure. Therefore, we can claim that our algorithm is able to find an optimal Bayesian network.

However, there are two main drawbacks to the existing algorithms: 1- As the number of variables increases (say more than 30) there must be a limit on the number of parents per node. 2- However by adopting different techniques such as dynamic programming the time to find an optimal network has reduced significantly, but still as the number of variablesgoes beyond ~15, the time to find an optimal network increases exponentially.

Exact structure Bayesian network structure learning with more than two parents per variables is a NP-Hard problem [2]. Therefore, we have to come up with a heuristic algorithm to find an optimal network. Leading researchers have adopted different techniques such as dynamic programming and linear programming relaxations to overcome many the NP-Hardness problem to some extent [3]-[6]. However, as the number of variables grows the problem shows its NP-Hard nature again.

The rest of paper is organized as follow. In section 2 we present the processes we have done on the data sets to make it ready for further analysis. In section 3 we present our analysis on the data sets using entropy. In section 4 we show how we used our algorithm to find an optimal Bayesian network structure for children handwriting data set. In section 5 we show how we did sampling on both Bayesian networks and present the result of inference on them. In section 6 we show that our algorithm is also able to construct a Markov network. In section 7 we show the result of our algorithm (with some setting of the parameters) with the existing state-of-the-art algorithms. In section 8 we discuss future works and extension of the presented algorithm.

## 2    The Data Sets

The data set belongs to the hand writing samples of the word 'and', written in cursive and printed, of children in grades 1 to 4 for an elementary school in Minnesota.

### 2.1    Inconsistency in the Data Sets

Some feature values are missing and denoted by 99 or -1: 99 and -1 meaning inconsistent and unassigned respectively. Therefore, we need to apply some preprocessing on the data in order to use the data to construct a Bayesian network and do inference tasks on it. The value -1 indicates that the value of the specific feature is unassigned, and the value -99 indicates that the feature is inconsistent for that particular data point. Also, some of the data points have more than 6 of their feature values unassigned/inconsistent and therefore we remove those data points from the data sets as they could be considered as noisy data points. However, there are still data points having missing/inconsistent feature values we first compute the probability of each discrete value for each of the features for a grade and then generate samples from the distribution to replace the missing values for that specific feature. A more clever way of replacing the missing/inconsistent feature values for a specific data point could be to consider take the sum of differences of the not missing values with the corresponding values in an ideal pattern, e.g. Zaner-Bloster copy book[1]. Similarly, for the data rows not missing those values we can compute the sum of differences with the ideal pattern and store each data point's sum of differences from the ideal pattern. Finally, by finding the data points having the closest difference with the data point having missed feature values we can replace the missing/inconsistent feature values with the ones from the non-corrupted data points having the closes difference with the data point having missing/inconsistent feature values. In addition, since more than 95 percent of the values corresponding to the feature and are either 99 or -1 we removed the corresponding columns in the data set for these two features. Finally, we will have the number of features reduced to 11.

After applying the above process on the data, we will add an auxiliary feature as the 12th feature storing the discrete value corresponding to each grade, e.g. the added feature for the 4th grade cursive hand writings will have the value 4. Also, there will be less than 50 data points left for the unclassified handwritings. Therefore, we do not take them into account for further analysis and structural learning of the Bayesian network.

## 3  Analysis on Data Sets

Statistical tools such as Entropy may reveal many important aspects about the features, their importance and relationships. Below, is the summarization of the results.

### 3.1  Entropy

Shannon's Entropy measures the uncertainty in the data. We compute entropy for the data sets using two approaches: 1- Computing the entropy for each grade of the same hand writing type separately. 2- Computing the entropy for combined data sets of cursive and printed hand writing.

#### 3.1.1 Entropy for Each Grade and Hand Writing

Having high entropy for features makes the discrimination process hard. As you can see in the figure 1, the cursive hand writings features for both grades have higher entropy than the features of printed hand writings. This is an indicative that it could be harder to do prediction tasks for cursive hand writings or it might mean that the set of extracted features for cursive hand writings are not the appropriate set of features leading to have high uncertainty and scarcity in the data. However, as shown in the figure 2 we can see that the entropy decreases going from each grade to a higher grades. This means that children's hand writings are becoming similar to each other or at least their writing patterns are becoming similar to each other, lowering the entropy. However, the feature No. 12 (the grade feature) has an entropy of 0. This is because the value for this feature is the same for each grade, e.g. it has a value of 3 for the 3rd grade and so on, and the uncertainty is zero.

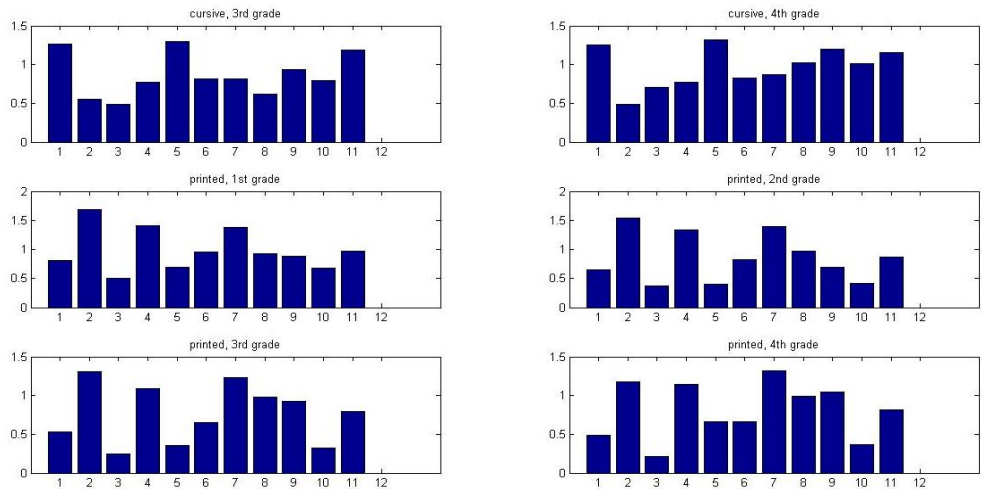

Figure 1: Entropy for each grade and hand writing

### 3.1.2 Difference of entropy measures between grades

After computing the entropy measures, we take the difference of the entropy measures for each grade to see how the entropy changes over time. The results are depicted in the figure 1. Looking at the differences, we notice that entropy keeps decreasing from lower grades to higher grades, in general. For instance, the entropy difference of printed hand writings for the children in 1st grade with 3rd and 4th grade shows a significant decrease in overall. This implies that the pattern of children's handwriting is becoming similar, either better or worse, over time leading to have less uncertainty in the features. However, for higher grades the difference of entropy measures do not change significantly. This may imply that after grades 3 or four the hand writings do not change significantly, and children have a steady way of writing. Also, entropy for the features 8 and 9 never decreases over time (except for the entropy difference between grades 1 and 2 which is negligible). This implies that these two features are the features that have an important impact on forming children hand writing and can play an important role in distinguishing different hand writings. However, we will check this after constructing two Bayesian network models: one having these two variables and one not.

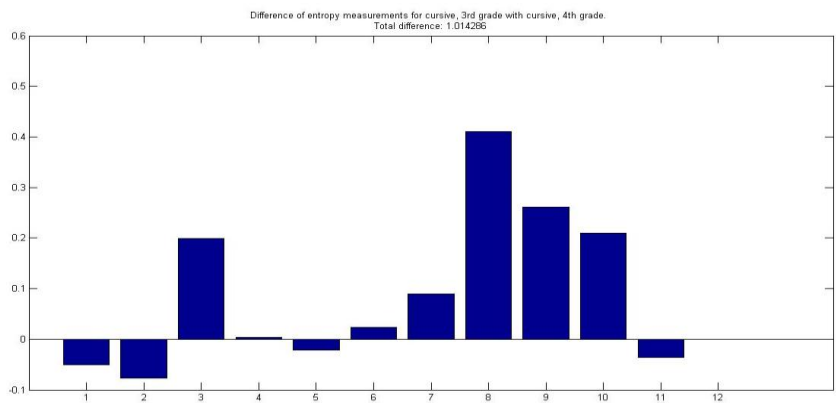

Figure 2: Difference of entropy of cursive hand writing for grades 3 and 4

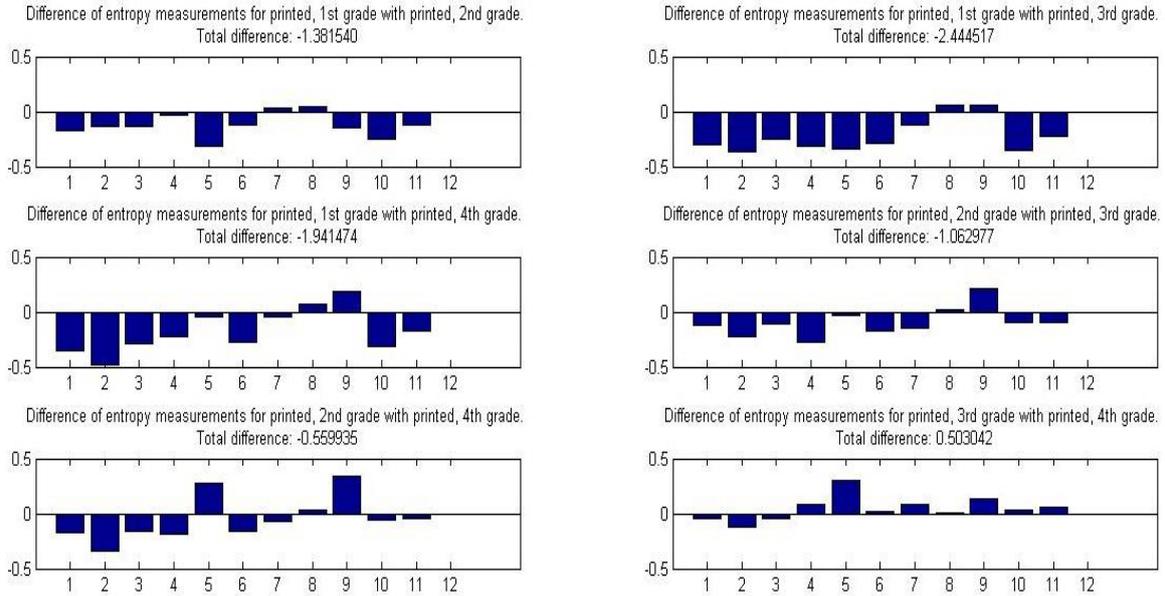

Figure 3: Difference of entropy for printed hand writing for grades 1 to 4

## 3.2 Rarity in the handwritings

We believe there are four approaches to determine rare hand writings in the data sets: (a) calculating the distance from an ideal pattern such as Zaner-Bloser as done in [1]. (b) Calculating the distance from the mean of the distributions and choosing the data points being lower or higher than a threshold from the mean. (c) Summing up the number of data points that are rarely represented in the data set (d) Projecting the data into 2-D space and selecting the data points being further from other data points in the 2-D space

We believe the last method is the best way of finding rare data points in the data since the first and second method are too naïve and the third method doesn't work well since . However, since the number of data points is not adequate the variance of error in finding rare hand writings could be large.

## 4 Modeling the data sets in Bayesian networks

We modeled the data sets using a novel algorithm we devised not any available toolbox. The algorithm uses decomposable scoring functions such as Bayesian Information Criterion (BIC), Akaike Information Criterion (AIC) or different variants of Bayesian Dirichlet (BD) to determine parents of each variable. The scoring function used for learning the final Bayesian network structure on which we will do inference for both cursive and printed data sets shown in figures 4 and 5 is AIC. However, we used all of the three scoring functions to construct a Bayesian network as shown in the pictures below.

The reason that we chose AIC over BIC is that although both scoring functions assume that the number of data points is infinite, but AIC penalizes the likelihood less than BIC does. Therefore, BIC helps capture strong dependencies among the variables which may lead to a very sparse network in case of having a data set with low number of data points. So, AIC is more suitable than BIC for this particular data set of children handwriting since the data set has a low number of data points. We could also use the network yielded by BDe scoring function for inference.

The BIC/AIC scoring function is as follow:

$$score_D(G) = max_\theta L_{G,D}(\theta) - t(G).w$$

Where $L_{G,D}(\theta) = N_{ijk} \log \frac{N_{ijk}}{\sum_{k'} N_{ijk'}}$ is the likelihood term, $t_i(G) = \sum_1^n (q_i.(r_i - 1))$ which represents the number of parameters in a graph with a sink node $i$, and w=1 for AIC and w=$\frac{\log N}{2}$ for BIC.

The BD scoring function is as follow:

$$score_i(pa_i) = \sum_{j=1}^{r_{pa_i}} (\log(\frac{\tau(a_{ij})}{\tau(a_{ij} + n_{ij})}) + \sum_{k=1}^{r_i} \log \frac{\tau(a_{ijk} + n_{ijk})}{\tau(a_{ijk})}$$

Where $\tau$ is the gamma function and $a_{ij}$ is a hyper-parameters.

### 4.1    The algorithm to learn Bayesian networks

The algorithm starts by setting the number of parents of each variable to be zero. Then the score for each variable is computed. On the next iteration the algorithm increases the number of parents to 1 and computes the score for each variable having one parent. Then the highest score and the parent yielding that score are stored for each of the variables.

On the next step the scores for the time that variables had no parent and the time that they had one parent are compared and the variables having a higher score (at least 5% higher) will be marked as the variables that still have the potential of adding parents to. Therefore, on the next iteration when the number of parents increases to 2 (or more), only those variables that are allowed to have more parents will make it to the next step and the algorithm does not compute any score for the new parent configurations of the variables marked as low-potential.

By setting the greediness level of the algorithm the running time will change from very short to long. The algorithm will find the most important dependencies between the variables in case of setting the parameters to the highest greedy level. These parameters have made our algorithm a flexible algorithm.

Here we briefly go over some of the parameters of the algorithm. The algorithm has about 7 parameters with which we can set a trade-off between accuracy and time. As an instance, having some simple assumptions such as not adding more parents to the variables marked as low-potential, reduces both the search space and computational cost significantly. Since computing the score includes creating the contingency table and then creating contingency frequency table by marginalizing over the child node's data vector. By experiments we noticed that if we focus on limiting the number of times that the algorithm must compute the score given a new parent configuration, the running time reduces significantly.

However having this bound will reduce the number of computations significantly, but to keep the running time very short for networks having a high number of variables, say more than 20, we need to add more bounds to reduce both the number of computations and move the search space to the optimal search space.

Two other important limitations we have in the algorithm are restricting the nodes to have the previously known parents in their new parent configurations which we call Parent Fidelity and the other limitation is using a simple grading method for each node and reduce the score whenever none of the node's new parent configuration yield a higher score.

While examining different methods we noticed that if we only use the previously known parents of a node in the subsequent parent configurations we will get almost exactly the same result compared to the time the algorithm searches on the completely unlimited space. By adding this limitation to the algorithm we reduce the search space significantly and move in the optimal search space with some error (less than 1% according to our experiments).

Also, if the score of the node having a new parent in its parent set is lower than the previously known score of the node, the algorithm reduces the node's grade by one. Since the grade for each node was set to 2 initially, the node will only have one more chance to have a new parent in its parent. Then after adding another parent to the node's parent set if the score is still lower than the last lowest score of the node the grade will set to 0 and the algorithm will never check new parent configurations for that node again.

However the algorithm is able to find the optimal Bayesian network without setting limit for the number of parents per node, we can still restrict the number of parents that we think is adequate for the algorithm.

Another parameter of the algorithm is grade model accuracy. The higher the model accuracy is set, the longer it takes the algorithm to find an optimal network matching the accuracy level. Note that the accuracy level is in percentage and will be taken into account for each node. Therefore, it is a rough estimate of the overall accuracy of the desired network.

There are also some other parameters that can highly impact the quality of the network, but we do not go over all the details of the algorithm here since we want to publish the algorithm in a conference and we can go over the details in that paper.

Below you can see a very abstract pseuducode version of the algorithm:

```
Bayesian Network Algorithm
Input: (dataset, parameters)
Output: Directed Acyclic Graph
Set parentsNum = 1
Set grade(1 * N) = 2
Set parentScores(N * N) = -inf
Set parentSets(1 * N) = {∅ }
Set addParentsToNodes(1 * N) = 1
Set numOfPa = 1
parentScores = score(dataset, 0)

Repeat
        applyRestrictions(parameters)
        PaConfigs = nchoosek(1:N, currentNumberofPa)
        For j=1 to sum(addParentsToNodes)
                PaConfigs = cleanPaConfig(greediness_parameters, PaConfigs)
                rows = numOfRows(paConfigs)
                For k=1 to rows
                        dataVectors = collect(dataset, paConfig(k))
                        newScore = score(dataVectors, paConfig)
                        If newScore > parentScores(numOfPa, j)
                                parentScores(numOfPa, j) = newScore
                End for k

        End for j
        grade = setGrade(numOfPa, parentScores)
Increment numOfPa by 1
Until (sum(addParentsToNodes) > 0)
```

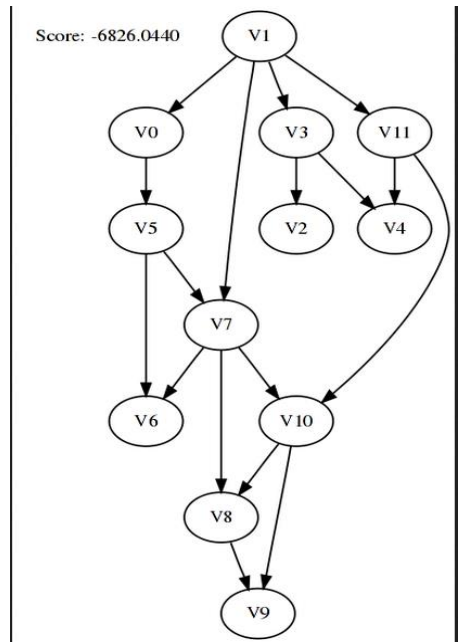

Figure 4: The generated network for printed hand writing using AIC

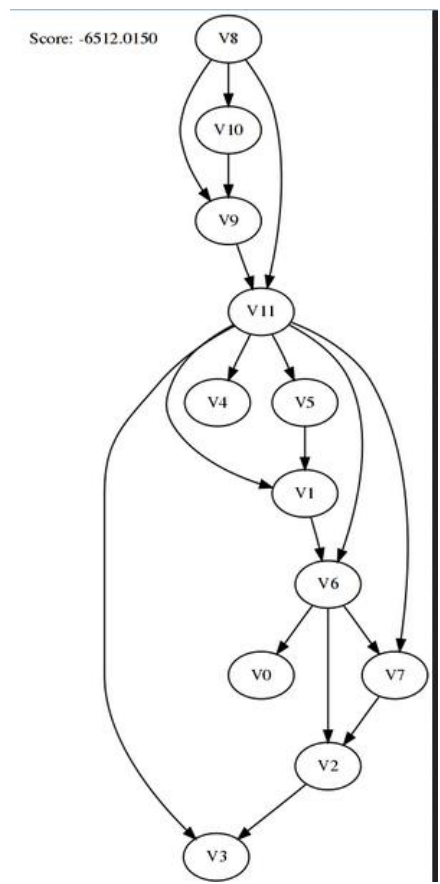

Figure 5: Constructed network for cursive hand writing using AIC

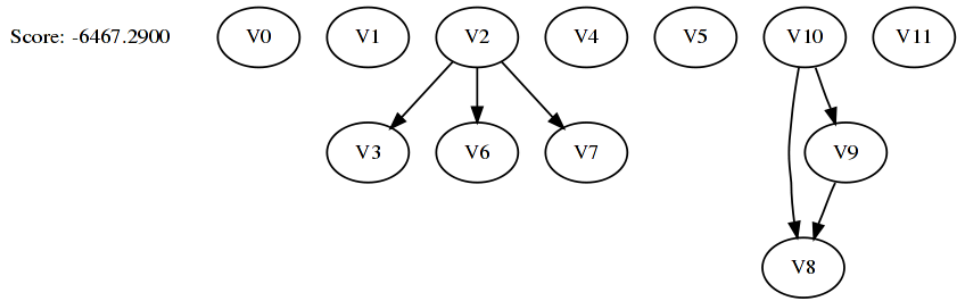

Figure 6: Constructed network for cursive hand writing using BIC

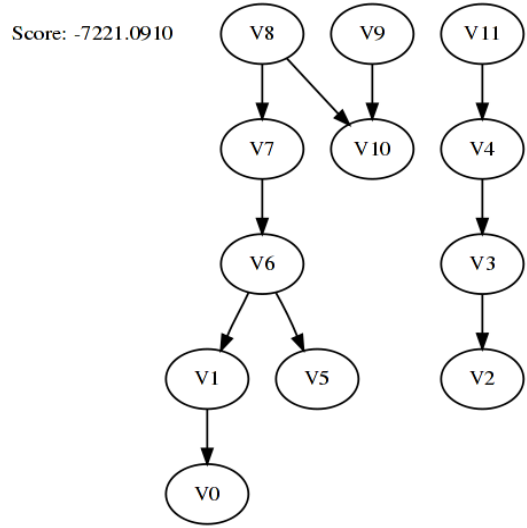

Figure 7: Constructed network for printed hand writing using BIC

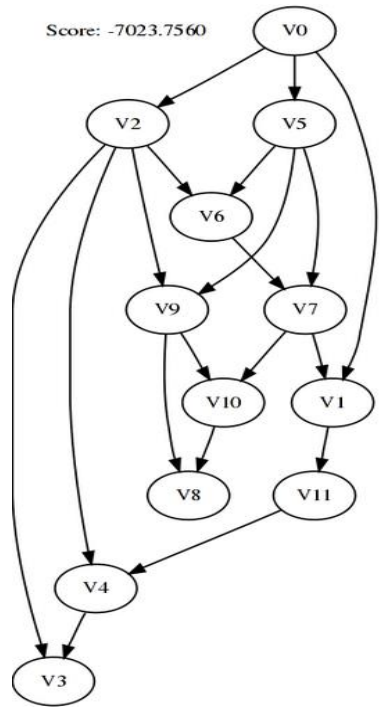

Figure 8: Constructed network for printed hand writing using BDe with ESS = 20

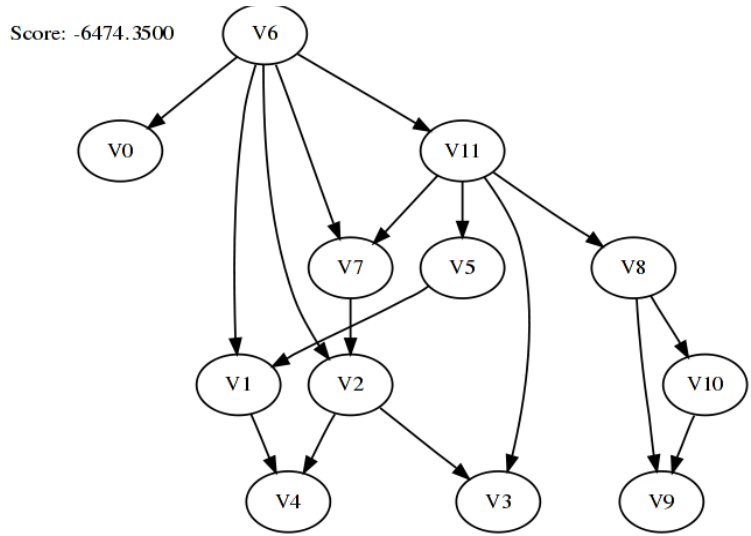

Figure 9: Constructed network for cursive hand writing using BDe with ESS = 20

We may note that we used toolbox provided by [5] to verify our results.

## 5 Inference

After learning the structure of the network, we store the conditional probability tables (CPDs) for the variables and run some queries to do inference on the model. To do topological inference we first need to extract the order of the network and then run the inference algorithm on it.

Since variable 11 represents the grade of the children, we will ask most of the queries on this variable.

### 5.1 Forward sampling

After having the topological order of the variables we run the sampling algorithm on the Bayesian networks we have. The algorithm to do the sampling is as follow (borrowed from Daphne Koller's book)

```
Algorithm 12.1 Forward Sampling in a Bayesian network
    Procedure Forward-Sample (
        B    // Bayesian network over X
    )
1       Let X_1, ..., X_n be a topological ordering of X
2       for i = 1, ..., n
3           u_i ← x⟨Pa_{X_i}⟩    // Assignment to Pa_{X_i} in x_1, ..., x_{i-1}
4           Sample x_i from P(X_i | u_i)
5       return (x_1, ..., x_n)
```

Since we want to have a minimum guaranteed accuracy for the generated samples we use the Hoeffding bound theorem and set the epsilon to 0.015 and set the value of delta to 0.985 for when we do not have a query.

$$M \geq \frac{\log\left(\frac{2}{0.985}\right)}{2 * 0.015}$$

Since we want to run some inference on the data set we have to use the following Hoeffding bound theorem:

$$M \geq 3 * \frac{\log\left(\frac{2}{0.985}\right)}{P(Y=y) * 2 * 0.015}$$

Now we are ready to do some inference on the network. Note that F12 is the $12^{th}$ feature of the data representing grade of students. It takes the values 1-4. For conditional probabilities we used the rejection sampling algorithm.

| Handwriting type | Query | Probability |
| --- | --- | --- |
| Cursive | P(F12=4) | 0.4682 |
| Cursive | P(F12=4 \| F1=2, F5=1, F9=2) | 0.5833 |
| Cursive | P(F12=3 \| F1=0, F3=1) | 0.5521 |
| Printed | P(F12=1 \| F4=1, F8=0) | 0.0828 |
| Printed | P(F1=2 \| F12=2, F9=0, F7=1) | 0.035 |

Table 1: Inference Results

## 6     Markov Networks

Since our algorithm is able to capture the dependencies between the variables we can use those dependencies to construct a Markov network.

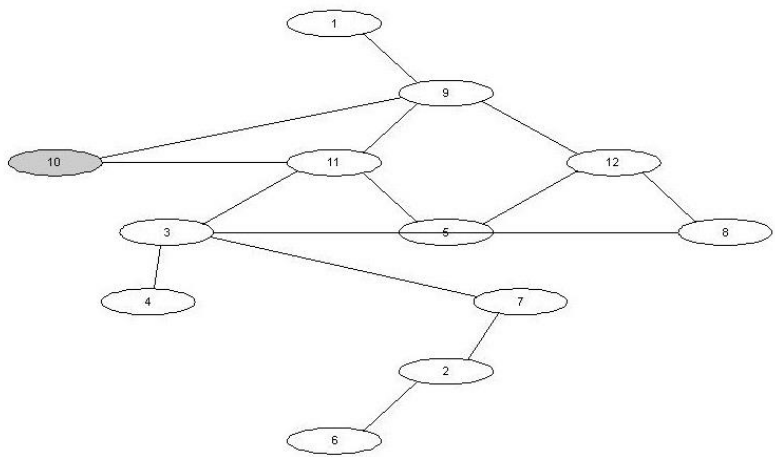

Figure 10: Constructed markov network for cursive hand writing

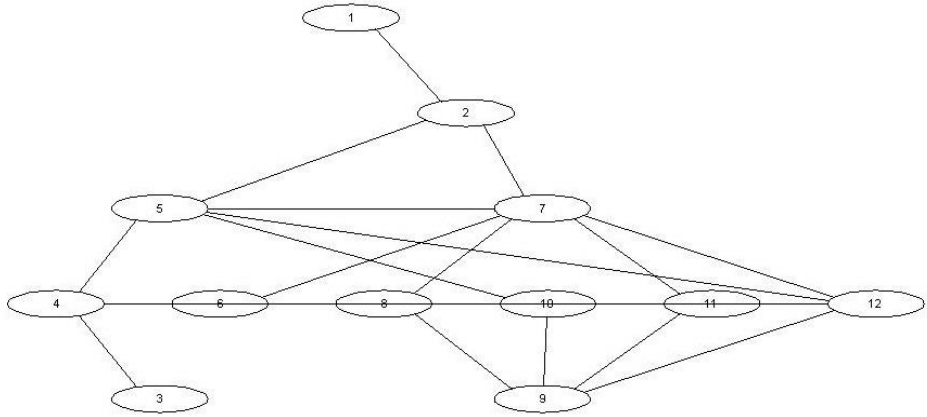

Figure 11: Constructed Markov network for printed hand writing

# 7    Results

Here we compare the results of our algorithm with [3] and [5]. Since we set the parameters of the algorithm in a way that it achieves the highest accuracy, the time of the algorithm increases. However, we have some ideas that if implemented that can reduce the running time significantly:

| Dataset | Number of Variables | **B&B** | | **DP** | | **Ours** | |
|---|---|---|---|---|---|---|---|
| | | BIC Score | Time | BIC Score | Time | BIC Score | Time |
| Adult | 15 | -286902.8 | ~200s | -286902.8 | 0.77 | -274129 (better) | 37s |
| Letter | 17 | -173716.2 | 574.1 | -163293.228 | 22.8 | -189688 (worse) | 37s |

# 7    Discussions and Future Work

We are currently working on the algorithm to fine-tune its parameters, try out new methods and add parameters to it to make it more flexible or accurate. We will also try applying optimization algorithms to this problem in a future work.